\pdfoutput=1

\documentclass[11pt]{article}

\usepackage[final]{acl}

\usepackage{times}
\usepackage{latexsym}

\usepackage[T1]{fontenc}

\usepackage[utf8]{inputenc}

\usepackage{microtype}

\usepackage{inconsolata}

\usepackage{graphicx}

\usepackage{subfigure}
\usepackage{amsmath,amsthm}
\usepackage{amssymb}
\usepackage{changes}
\definechangesauthor[name=wenyuchen]{wc}
\definechangesauthor[name=wenbo,color=red]{wb}

\usepackage[english]{babel}
\usepackage{mathtools}
\newtheorem{theorem}{Theorem}[section]
\newtheorem{lemma}[theorem]{Lemma}

\usepackage{multirow}
\usepackage{paracol}
\usepackage{booktabs} 
\usepackage{authblk}

\title{Beyond the Singular: Revealing the Value of Multiple Generations in Benchmark Evaluation}

\author{
 \textbf{Wenbo Zhang\textsuperscript{1 $*$}},
 \textbf{Hengrui Cai\textsuperscript{1 $\dag$}},
 \textbf{Wenyu Chen\textsuperscript{2 $\dag$}}
\\
 \textsuperscript{1}University of California Irvine,
 \textsuperscript{2}Meta, Central Applied Science\\
  \texttt{\{wenbz13,hengrc1\}@uci.edu}, \texttt{wenyuchen@meta.com}
}

\begin{document}
\maketitle

\begingroup
\renewcommand{\thefootnote}{}
\footnotetext{\textsuperscript{$*$} Work done during internship at Meta}
\footnotetext{\textsuperscript{$\dag$} Co-correspondence}
\endgroup

\begin{abstract}
Large language models (LLMs) have demonstrated significant utility in real-world applications, exhibiting impressive capabilities in natural language processing and understanding. Benchmark evaluations are crucial for assessing the capabilities of LLMs as they can provide a comprehensive assessment of their strengths and weaknesses. However, current evaluation methods often overlook the inherent randomness of LLMs by employing deterministic generation strategies or relying on a single random sample, resulting in unaccounted sampling variance and unreliable benchmark score estimates. In this paper, we propose a hierarchical statistical model that provides a more comprehensive representation of the benchmarking process by incorporating both benchmark characteristics and LLM randomness. We show that leveraging multiple generations improves the accuracy of estimating the benchmark score and reduces variance. Multiple generations also allow us to define $\mathbb P\left(\text{correct}\right)$, a prompt-level difficulty score based on correct ratios, providing fine-grained insights into individual prompts. Additionally, we create a data map that visualizes difficulty and semantics of prompts, enabling error detection and quality control in benchmark construction.

\end{abstract}

\section{Introduction}
In recent years, advanced large language models have demonstrated remarkable versatility across a wide range of tasks and domains, with their development continuing to accelerate. To effectively track their progress, numerous generative benchmark datasets have been curated to assess both their general and specialized capabilities. 


There are two primary ways for generating responses from large language models (LLMs): greedy decoding and random sampling \citep{holtzman2019curious}. Greedy decoding selects the next token with the highest probability, resulting in a deterministic output. In contrast, random sampling, such as nucleus sampling \citep{holtzman2019curious}, incorporates randomness during decoding by sampling a token at each step based on a probability distribution. This approach leads to non-deterministic output. Current LLM benchmarks typically employ one of these methods; for instance, LiveBench \citep{white2024livebench} WildBench \citep{lin2024wildbench} and OpenLLM leaderboard \citep{open-llm-leaderboard-v1} use greedy decoding, while TrustLLM \citep{huang2024trustllm},  MT Bench \citep{zheng2023judging} and Alpaca Eval \citep{alpaca_eval} employ a non-deterministic sampling configuration. During evaluations, LLMs generate a single response for each prompt in the benchmark, and the correctness of these responses is determined by comparing them to the ground truth answers. The final benchmark score is then calculated as the average of these individual scores.


However, this presents challenges within the current generative-evaluation paradigm. Firstly, deterministic generation does not align with the real-world application of LLMs, where randomness is inherent. This misalignment can lead to biased estimations of LLM performance. Even with random generation, relying on a single generation can result in significant variance in benchmark scores, particularly when the sample size is small. Furthermore, a single generation is not sufficiently informative for individual prompts, as it cannot address prompt-level questions such as, "Which question is more challenging?" This limitation creates obstacles to understanding the overall composition of the benchmark data.

In this paper, we regard the benchmark as an estimation problem characterized by a statistical model and highlight the significance of incorporating multiple random generations in a principled way. We theoretically demonstrate that increasing the number of generations decreases the variance in benchmark score estimation. Moreover, by leveraging multiple samples, we introduce a fine-grained difficulty metric, $\mathbb P\left(\text{correct}\right)$, derived from the inherent latent parameters of our statistical model, to quantify the difficulty of individual prompts. This enables comparisons across different prompts. Additionally, we demonstrate that mislabeled or ambiguous prompts can be effectively detected using multiple generations, highlighting its potential as a tool in benchmark construction.


\section{Benchmarking Procedure is a Hierarchical Model}

In this section, we show that the benchmark is an estimation problem. Without loss of generality, we consider random sampling as the generation strategy where each token is randomly sampled from a token distribution conditional on previously generated tokens. We also assume the correctness of generations can be obtained using a judgment function, which can be accomplished either by comparing the response with ground truth or by determining whether it passes unit tests. 


Given an LLM parameterized by parameters $\theta$, including both model parameters and sampling parameters, for example temperature $T$ and top $P$, etc.), and a benchmark dataset $\mathcal{D}=\{x_i\}_{i=1}^n$, we can define difficulty of the $i$-th prompt with respect to the LLM as a random variable drawn from the unknown benchmark difficulty distribution $\mathbb{P}(\mu, \sigma;\theta)$, with mean $\mu$ and standard deviation $\sigma$. Without loss of generality, with $k$ generations per prompt, we can then regard the benchmarking procedure as a hierarchical model as follows:
\begin{equation}
\begin{gathered}
p_i \sim \mathbb{P}(\mu, \sigma;\theta) \quad \text{for } i=1,\cdots,n,\\
y_{i,j}  \sim \text{Bernoulli}(p_i) \quad \text{for } j=1,\cdots,k,
\end{gathered}
\label{stat model}
\end{equation}
where prompt difficulty $p_i$ is sampled from $\mathbb{P}(\mu, \sigma;\theta)$ and $p_i$ represents the probability that the LLM can correctly answer the $i$-th prompt., i.e., $\mathbb P\left(\text{A generated answer to $i$-th prompt is correct}\right)=p_i$. This represents a latent difficulty of prompts, We denote the he $k$-th generation of the $i-$th prompt as $z_{i,j}$ and then $y_{i,j}$ is the correctness indicator for it, where $y_{i,j}=1$ if it’s correct otherwise $y_{i,k}=0$. 

 Here both benchmark distribution $\mathbb{P}(\mu, \sigma;\mathcal{D})$ and $p_i$ are unknown needs to be estimated with $\{y_{i,j}\}_{j=1}^k$ for $i=1, \cdots, n$.

To estimate $p_i$ and $\mu$, we can use a straight forward method of moment estimators $\hat p_i = \frac{\sum_{j=1}^k y_{i,j}}{k}$, $\hat \mu = \frac{\sum_{i=1}^n \hat p_{i}}{n}=\frac{\sum_{i=1}^n\sum_{j=1}^k y_{i,j}}{nk}$. We observe that a widely used item response theory \citep{polo2024tinybenchmarks,madaan2024quantifying,ding2024easy2hard}, employed to model the difficulty of prompts, represents a specific parametrization of $\mathbb{P}(\mu, \sigma;\mathcal{D})$. Further elaboration on this can be found in Appendix \ref{irt section}.

Note that, when $k=1$, the benchmark score computed based on a single random generation is an estimation of $\mu$, which only utilizes a single generation which leads to a large variance. We can show this by explicitly calculating the variance of our estimators.

\begin{lemma}
    Given the  hierarchical model in (\ref{stat model}) and the moment estimators $\hat \mu =\frac{\sum_{i=1}^n\sum_{j=1}^k y_{i,j}}{nk}.$
    Then $\hat \mu$ is an unbiased estimator for $\mu$ and its variance equals:
    \begin{equation}
\text{Var}(\hat\mu)=\underbrace{\frac{1}{nk}\left(\mu-\mu^2-\sigma^2\right)}_{\text{Withth-prompt Variance}}+\underbrace{\frac{1}{n}\sigma^2}_{\text{Between-prompt Variance}}.
\label{var}
    \end{equation}
    \label{lemma}
    \vspace{-0.5cm}
\end{lemma}

Here, $\text{Var}(\hat\mu)$ can be decomposed into within-prompt variance and between-prompt variance. Both terms decrease as the number of benchmark data $n$ increases. However, since benchmark data is typically fixed, we analyze the influence of sampling in terms of $k$. Within-prompt variance captures the randomness in sampling $y_{ij}$ conditional on the $i-$th prompt, and it can be effectively reduced by increasing the number of samples $k$, converging to $0$ as $k \to \infty$. The between-prompt variance term, on the other hand, captures the variability of prompt difficulty $p_i$ across groups, reflecting the randomness of difficulty distribution $\mathbb{P}(\mu, \sigma;\theta)$, and thus remains unaffected by $k$.

We can further plug in sample variance $\hat\sigma^2=\frac{1}{n-1} \sum_{i=1}^n(\hat p_{i}-\frac{\sum_{i=1}^n \hat p_i}{n})^2$ and $\hat\mu$ into (\ref{var}) to get $\widehat{\text{Var}(\hat \mu)}$. Finally, based on the central limit theorem, a $95\%$ confidence interval is: $\hat\mu \pm 1.96 \sqrt{\widehat{\text{Var}(\hat \mu)}}$.

\begin{figure}[!htbp]
    \centering
        \subfigure[MMLU-Pro]{\includegraphics[width=0.46\columnwidth]{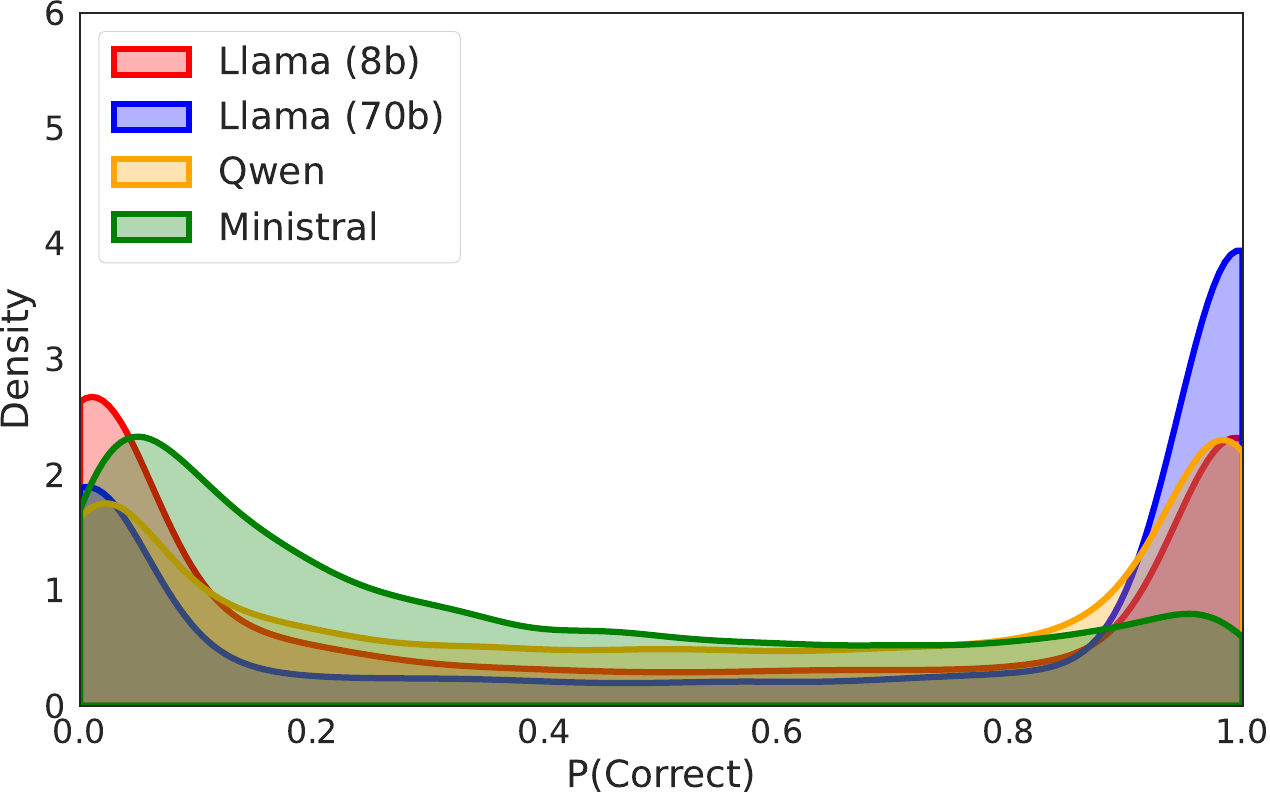}}
    \subfigure[GSM8K]{\includegraphics[width=0.46\columnwidth]{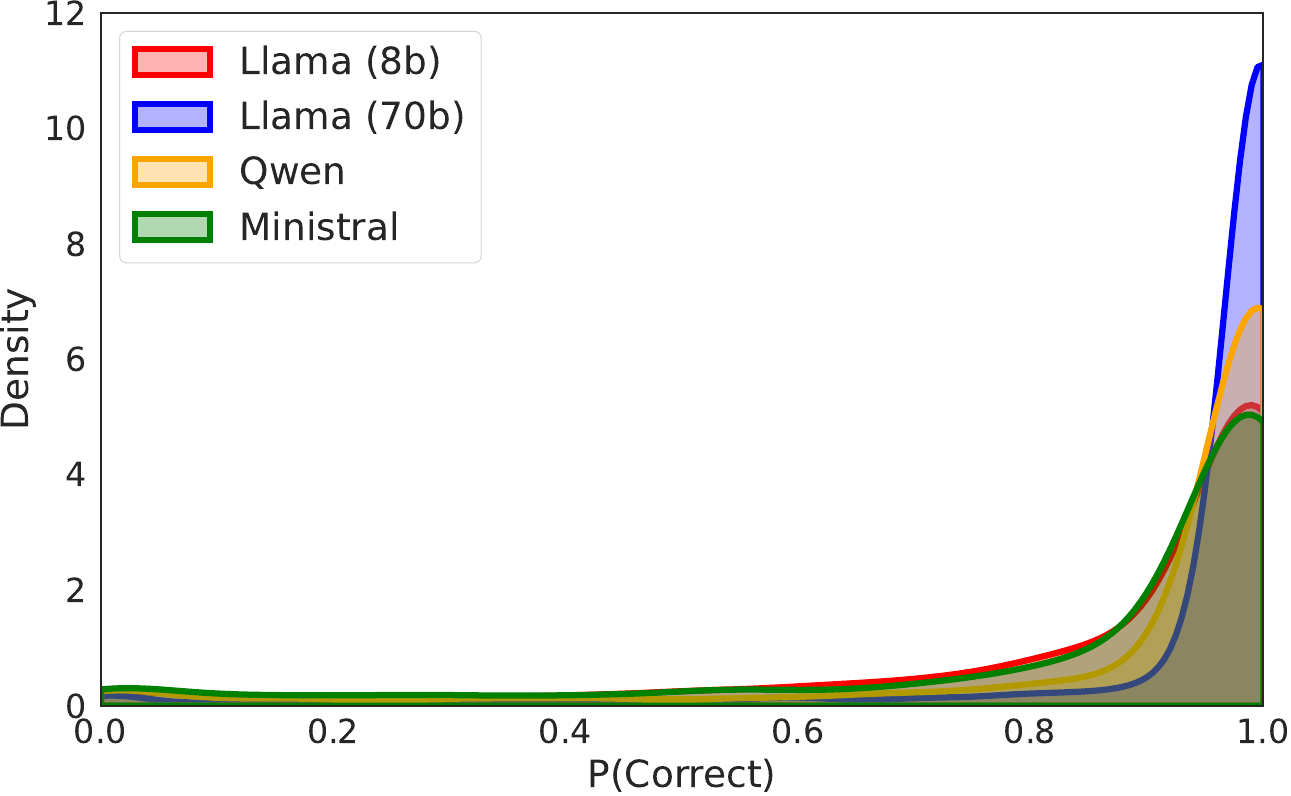}}\\
    \subfigure[IFEval]{\includegraphics[width=0.46\columnwidth]{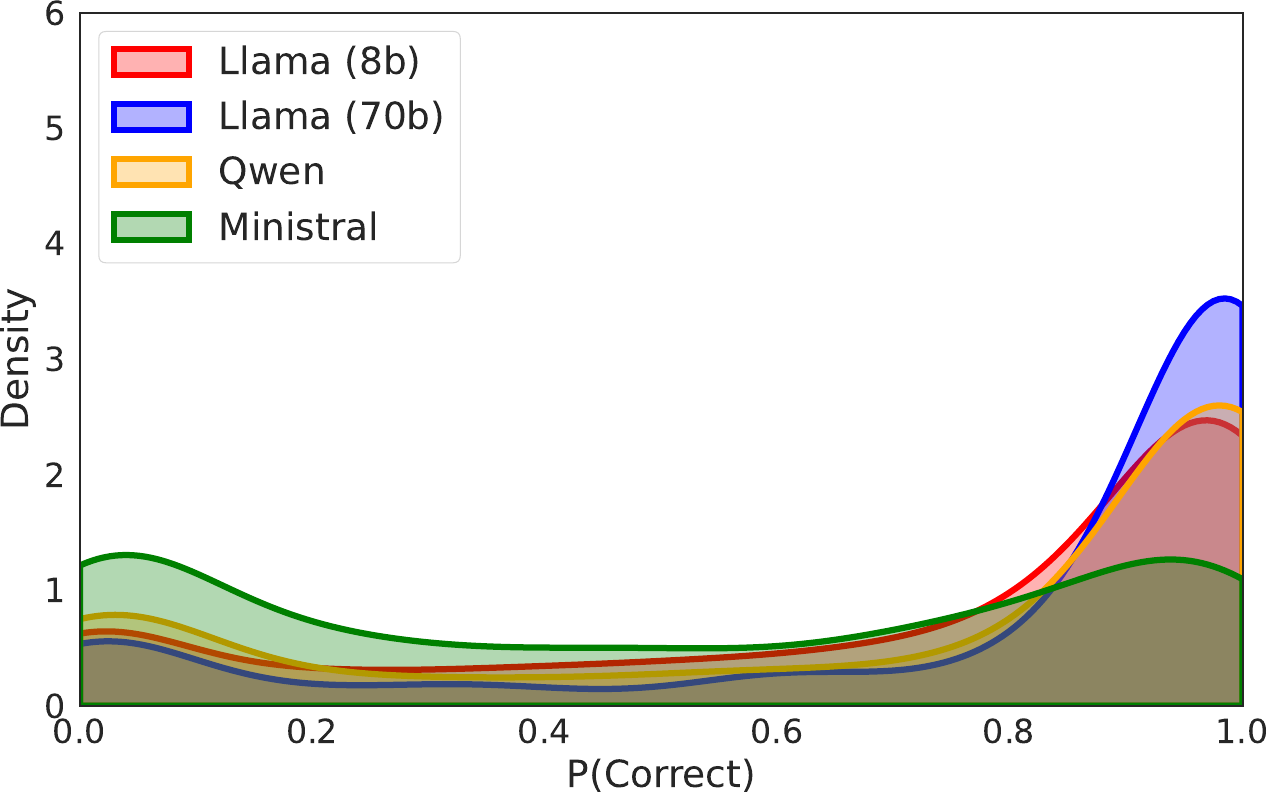}}
    \subfigure[MuSR]{\includegraphics[width=0.46\columnwidth]{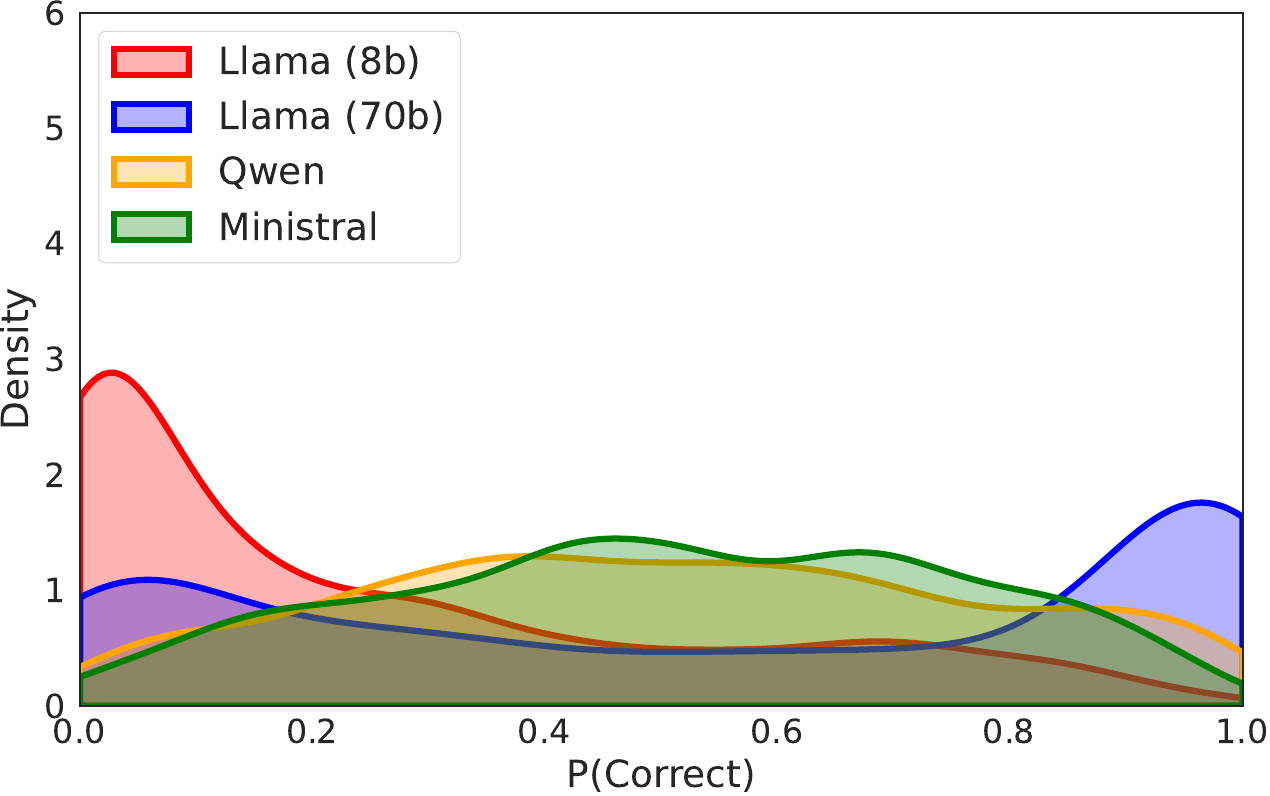}}
    \\
    \caption{Distribution of $\mathbb P\left(\text{correct}\right)$ of $4$ benchmarks.}
    \label{fig:dist}
    \vspace{-5mm}
\end{figure}
\subsection{Prompt Level Difficulty: $\mathbb P\left(\text{correct}\right)$}
Our goal is to develop a granular, quantifiable measure of prompt difficulty, enabling us to gain a deeper understanding of their relative complexities. By quantifying prompt difficulty at the individual level, we can address fundamental questions such as: `Which prompts are most challenging?' and `How do different prompts compare in terms of difficulty?' A fine-grained understanding of prompt difficulty will provide valuable insights into the strengths and weaknesses of language models, as well as the composition of benchmark datasets, ultimately informing the development of more effective models and evaluation frameworks. 

We refer to $\mathbb P\left(\text{correct}\right)=p_i$ in (\ref{stat model}) and its estimation $\widehat{ \mathbb{P}}\left(\text{correct}\right)=\hat p_i = \frac{\sum_{j=1}^k y_{i,j}}{k}$. When the number of generations $k$ increases, it will converge to the true $\mathbb P\left(\text{correct}\right)$ and therefore more fine-grained. The probability of correctness $p_i$ can be interpreted as a difficulty score at the prompt level: the higher the $p_i$, the easier the prompt since the language model has a higher probability of generating a correct response. We demonstrate the use of difficulty scores in the analysis section.




\section{Experiments}
\subsection{Experimental Setup}
\noindent \textbf{Benchmark.}
We choose multiple benchmarks which cover various capabilities of LLMs: MMLU-Pro \citep{wang2024mmlu}, GSM8K \citep{cobbe2021training}, MuSR \citep{sprague2023musr}, IFEval \citep{zhou2023instruction}. For MMLU-Pro, GSM8K, and MUSR, we use accuracy as the metric, while for IFEval, we utilize instance-level strict accuracy. More details of benchmarks are in Appendix \ref{bench details}.

\noindent \textbf{LLM and Setup.}
We utilize four widely-used open-source LLMs: Llama 3.1 (8B and 70B Instruct) \citep{dubey2024llama}, Qwen 2.5 (7B Instruct) \citep{qwen2}, and Ministral (8B Instruct) \citep{jiang2023mistral}\footnote{Ministral models and analysis on Ministral output were run only by some authors on academic research systems.}. We evaluate both greedy decoding and random sampling on these models, with the latter using a temperature of $0.7$ and top-p of $1.0$. For each prompt across all benchmarks, we generate $50$ samples ($k=50$) using a 0-shot chain-of-thought prompting strategy.

\begin{table*}[t]

\centering

\caption{Results on four benchmark datasets with four open source LLMs. "n" is the number of prompts, "Greedy" denotes greedy decoding, "Sample (k=50)" is the random sample with $50$ generations and "$\Delta$ ($k=1$)" denotes the performance gap between the best and worst run with $1$ generation. We include both benchmark score and SE.}

\resizebox{1\textwidth}{!}{

\begin{tabular}{lccccccc}

\toprule \multirow{2}{*}{\textbf{Benchmark}} & \multirow{2}{*}{\textbf{n}} & \multicolumn{3}{c}{\textbf{Llama 3.1 8b Instruct}} & \multicolumn{3}{c}{\textbf{Llama3.1 70b Instruct}}  \\
\cmidrule(lr){3-5} \cmidrule(lr){6-8} 
& & \textbf{Greedy} & \textbf{Sample} ($k=50$)& $\Delta(k=1)$  & \textbf{Greedy} & \textbf{Sample} ($k=50$) & $\Delta (k=1)$ \\

\hline MMLU-Pro  & $12,187$ & $46.2$ ($0.45$) & $46.1$ ($0.39$) & $10.0$ & $63.8$ ($0.44$)  & $63.4$ ($0.40$) & $3.9$ \\

 GSM8K & $1,319$ & $86.1$ ($0.95$) & $85.6$ ($0.68$) & $18.6$ & $95.6$ ($0.56$) & $95.3$ ($0.45$) & $4.8$ \\

 IFEval & $541$ & $74.5$ ($1.87$) & $71.1$ ($1.51$) & $8.3$ & $82.6$ ($1.64$) & $80.2$ ($1.42$)  & $5.9$ \\

 MuSR & $756$ & $24.8$ ($1.65$)  & $29.0$ ($1.00$) & $8.2$ & $56.3$ ($1.80$)  & $57.9$ ($1.40$) & $5.4$ \\
\midrule \multirow{2}{*}{\textbf{Benchmark}} & \multirow{2}{*}{\textbf{n}} & \multicolumn{3}{c}{\textbf{Qwen 2.5 7B Instruct}} & \multicolumn{3}{c}{\textbf{Ministral 8B Instruct}}  \\
\cmidrule(lr){3-5} \cmidrule(lr){6-8} 
& & \textbf{Greedy} & \textbf{Sample ($k=50$)}& $\Delta (k=1)$  & \textbf{Greedy} & \textbf{Sample} ($k=50$) & $\Delta (k=1)$ \\

\hline MMLU-Pro  & $12,187$ & $53.3$ ($0.45$) & $53.0$ ($0.36$) & $1.3$ & $39.7$ ($0.44$)  & $36.3$ ($0.29$) & $1.5$ \\

 GSM8K & $1,319$ & $90.2$ ($0.82$) & $90.2$ ($0.65$) & $2.3$ & $86.1$ ($0.95$) & $84.9$ ($0.73$) & $3.1$ \\

 IFEval & $541$ & $72.6$ ($1.92$) & $71.2$ ($1.64$) & $5.9$ & $51.4$ ($2.15$) & $49.8$ ($1.65$)  & $5.6$ \\

MuSR & $756$ & $49.2$ ($1.82$)  & $50.9$ ($0.98$) & $8.3$ & $49.7$ ($1.82$)  & $50.8$ ($0.91$) & $8.6$ \\
\bottomrule
\end{tabular}
}
\label{full table}
\end{table*}

\subsection{Main Results}
Results are shown in Figures \ref{fig:dist} and Table \ref{full table}. Key takeaways are summarized below.

\noindent \textbf{Distribution of $\mathbb P\left(\text{correct}\right)$ show diffuse density in challenging tasks, behaving like random samplers.}
For the distribution of $\mathbb P\left(\text{correct}\right)$, we define stable behavior as a density distribution with high concentrations near $0$ and $1$, and lower density in between. Conversely, a distribution with a high density between $0$ and $1$ indicates high randomness. As shown in Figure \ref{fig:dist}, when confronted with benchmarks that require strong reasoning skills (MMLU-Pro, IFEval, and MuSR), all models display a diffuse density distribution over the support $\left[0,1\right]$. This suggests that LLMs resemble random samplers when handling prompts requiring strong reasoning, underscoring the complexity and sensitivity of their reasoning processes. In contrast, the simpler task GSM8K display densities with more pronounced tails and reduced uncertainty. A plausible explanation is that GSM8K is easier and involves shorter reasoning lengths, which in turn decreases the likelihood of diverse reasoning paths emerging. Additionally, we observe that the Llama 70B model exhibits the most stable performance across all benchmarks, suggesting that larger models can provide more stable reasoning process.

\begin{figure}[!htbp]
    \centering
    \subfigure{\includegraphics[width=0.46\columnwidth]{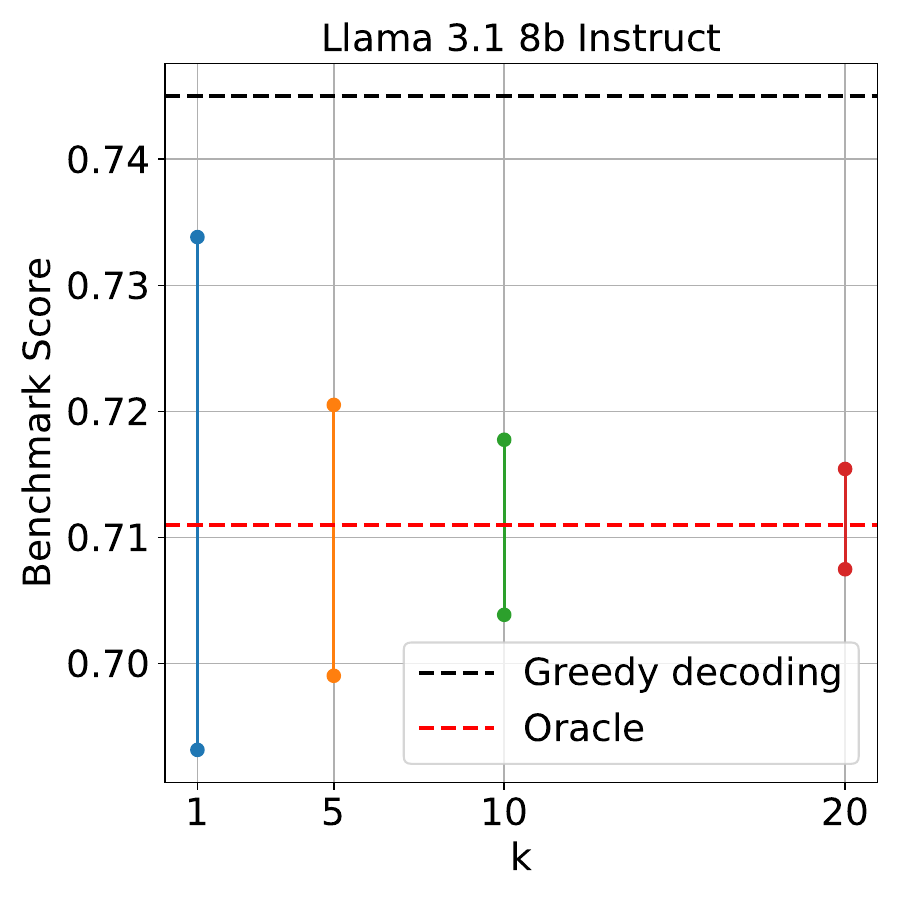}}
    \subfigure{\includegraphics[width=0.46\columnwidth]{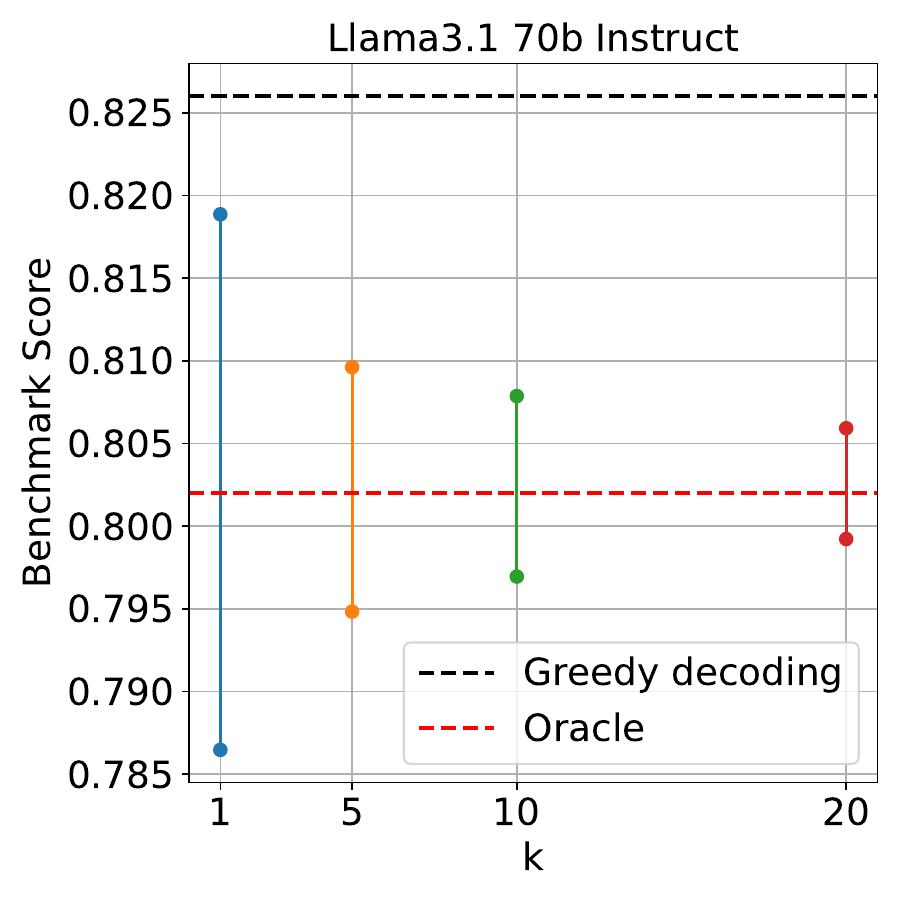}}\\
    \vspace{-0.5cm}
    \caption{Benchmark score of IFEval over different $k$. }
    \label{fig:ifeval over k}
    \vspace{-5mm}
\end{figure}

\noindent \textbf{Estimation differs noticeably between greedy decoding and random sampling, with a single random generation being unstable.}
Table \ref{full table} presents the benchmark scores, highlighting the performance differences between greedy decoding and random sampling. Notably, for GSM8K and MuSR, the absolute differences in benchmark score between these two methods for Llama3 8B are $3.4$ and $4.2$, respectively, indicating a relatively large performance gap. This discrepancy can also be observed in other models and datasets. Furthermore, we observe considerable variability with one generation, characterized by large values of $\Delta$($k=1$). This suggests that random sampling with limited generations is ineffective for benchmark evaluation, particularly for small datasets, aligning with our Lemma \ref{lemma}. 
We also investigate how sampling parameters influence the $\mathbb P\left(\text{correct}\right)$ distribution, and results are in Appendix \ref{vary T}. We further conduct a synthetic analysis to demonstrate the value of multiple generations. Using $k = 50$ as the oracle (i.e., the full set of generated samples), we evaluate $k = 1, 5, 10, 20$ over 1000 trials each by sampling with replacement. As shown in Fig. \ref{fig:ifeval over k}, increasing $k$ leads to narrower $95\%$ confidence intervals that coverage the true score. In contrast, greedy decoding exhibits a consistent performance gap, suggesting that even a modest number of sampled generations better approximates $\mathbb{P}(\text{correct})$ than greedy decoding.

\begin{figure}[htbp]
    \centering
    \includegraphics[width=\columnwidth]{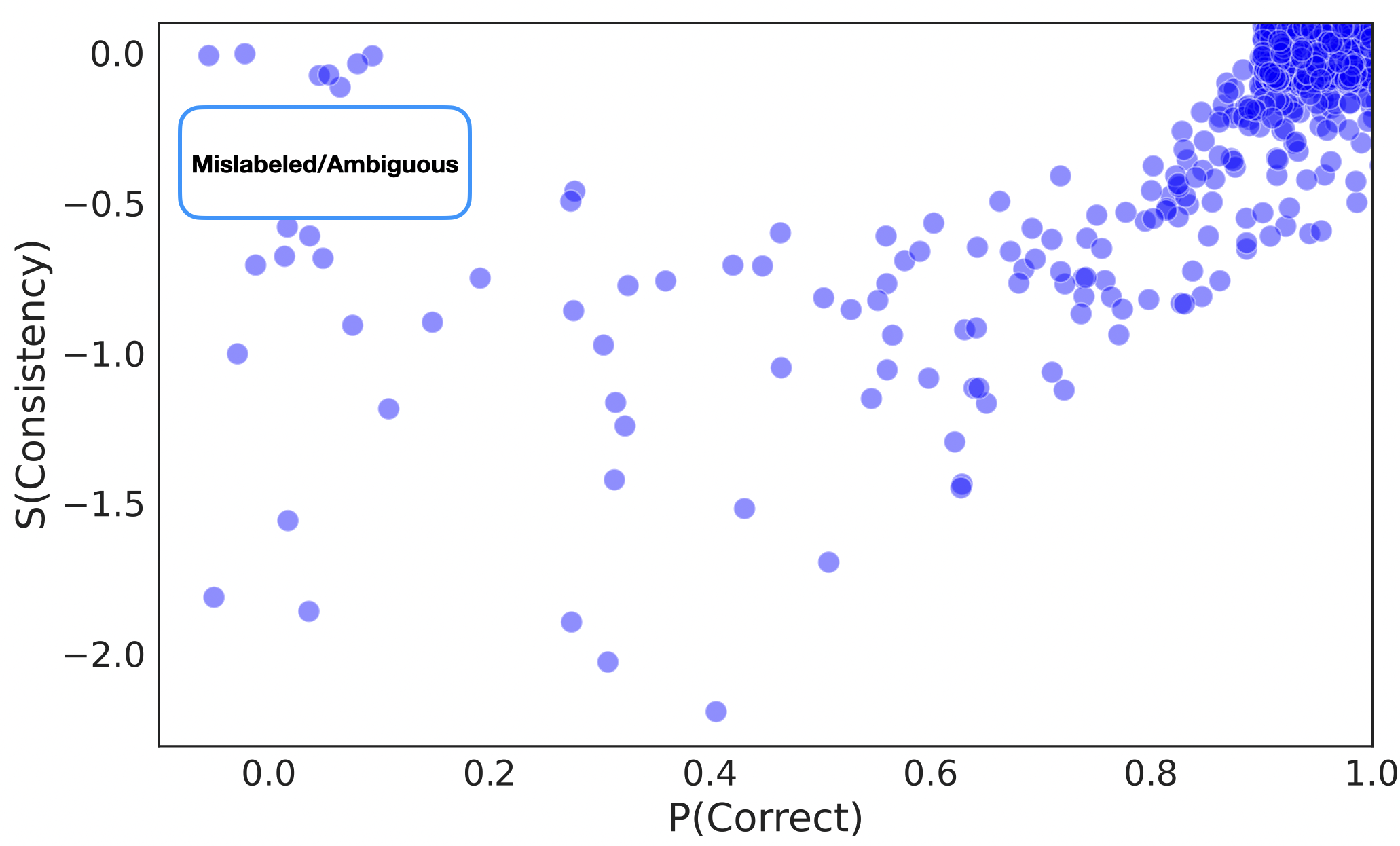}
    \caption{Data map for GSM8K with Llama 70b.}
    \label{fig:map}
\end{figure}

\noindent \textbf{Multiple generations can help detect labeling errors: a case study on GSM8K.}
Benchmark construction can involve label errors or ambiguous prompts, such as the approximately $5\%$ error rate in GSM8K. Manually cleaning large datasets is costly, but we found that using multiple generations from advanced LLMs can help identify mislabeled or ambiguous prompts. Based on multiple generations, we can create a data map to visualize $\mathbb{P}(\text{correct})$ against $\mathbb{S}(\text{consistency})$, which measures the semantic consistency of generations. Given a set of $k$ generations and clustering them into $C$ semantic sets, $\mathbb{S} \text { (consistency) } $ is defined as: $
\mathbb S\left(\text{consistency}\right) = \sum_{c=1}^C \text{Prop}_c\log \text{Prop}_c,
$
where $\text{Prop}_c$ measures the proportion of generations in group $c$ and its empirical estimator $\widehat{\text{Prop}}_c=\frac{\text{\# generations in set c}}{k}$. This can be seen as negative semantic set entropy; the larger, the more consistent. Semantic clusters in GSM8K can be derived from final answers and can be extended to more open-ended QA by embeddings or LLMs as judges.
We hypothesize that prompts with low $\mathbb P(\text{correct})$ and high $\mathbb{S}(\text{consistency})$ may be mislabeled or ambiguous due to contradicting with the self-consistency \citep{wang2022self}. Self-consistency \citep{wang2022self,mitchell2022enhancing} leverages the intuition that a challenging reasoning problem typically admits multiple reasoning paths leading to its unique correct answer. To verify our hypothesis, we utilize the data map of Llama3 70B for GSM8K and selected prompts with $\mathbb{P}(\text{correct}) \leq 0.1$ and $\mathbb{S}(\text{consistency}) \geq -0.8$, totaling $18$ prompts. After manually reviewing the selected prompts, we found that $44.4\%$ prompts were either mislabeled or ambiguous (having multiple valid interpretations of a question). Examples are shown in the Appendix Figure \ref{example}. Our results demonstrate the potential of data maps for dataset cleaning, extending prior work \citep{swayamdipta2020dataset} from classification to generative models. Notably, our approach only utilizes a single LLM and a simple semantic metric, underscoring future research opportunities to enhance accuracy through multiple models and improved semantic metrics.

\section{Related Work}

\subsection{LLM Benchmark Evaluation}
Recent benchmark evaluations have significantly enhanced our understanding of Large Language Models (LLMs) and have driven further advancements in the field. Notable benchmarks like MMLU \citep{hendrycks2020measuring}, HELM \citep{liang2022holistic}, and BIG-bench \citep{srivastava2022beyond} have expanded assessments to include language generation, general knowledge understanding, and complex reasoning. Several other benchmarks assess the trustworthiness of large language models (LLMs) \citep{wang2023decodingtrust, huang2024trustllm, zhang2024defining} in terms of safety, bias, privacy, and hallucination, etc. Leaderboards like the OpenLLM Leaderboard \citep{open-llm-leaderboard-v1} facilitate performance comparisons across LLMs by evaluating a range of tasks, each targeting different capabilities, to provide a comprehensive assessment of LLMs. However, most benchmark evaluations, even on leaderboards, rely on a single output per example, either greedy decoding or random sampling. \citet{song2024good} also examines the performance gap between the two types of generation strategies and highlights the importance of randomness. There is also concurrent work by \citet{miller2024adding} that mentions using multiple generations to reduce variance, but their contribution is primarily conceptual. In contrast, we provide both theoretical support and empirical results. Additionally, we propose several benefits of using multiple generations, such as difficulty quantification and mislabeled prompt detection, which distinguish our work from theirs.

\subsection{Prompt Difficulty in Benchmark}
Understanding prompt-level difficulty is crucial for analyzing benchmark composition and some benchmark datasets include difficulty scores for each prompt provided by humans. For example, the MATH dataset \citep{hendrycks2measuring} offers a variety of high-school-level problems with a broad five-level difficulty rating. Similarly, the GPQA dataset \citep{rein2023gpqa} contains graduate-level multiple-choice questions rated on a 4-point scale by two experts. Recent studies \citep{ding2024easy2hard,polotinybenchmarks} also attempted to estimate difficulty scores of individual prompts using item response theory \citep{cai2016item,natesan2016bayesian} or Glicko-2 \citep{glickman2012example}, based on offline evaluation results from a pool of large language models (LLMs) or human participants. This approach seeks to provide an objective difficulty score by encompassing a diverse range of testers, including both humans and LLMs. However, this can lead to misalignment when focusing solely on a target LLM. A question that is easy for one model might be difficult for others, highlighting the inherently subjective nature of difficulty \citep{desender2017subjective}. Therefore, it is more relevant to consider the subjective difficulty specific to the target LLM.

\section{Conclusion}
In this paper, we investigate the value of multiple generations in LLM benchmark evaluation. By leveraging a hierarchical model, we show that multiple generations help quantify prompt difficulty, reduce variance, and detect labeling errors, making evaluations more robust and informative.

\section*{Limitations}
While using multiple generations in benchmark evaluation is promising, it demands more computational resources during inference time. Future research could explore the minimal number of generations required for robust evaluation, potentially reducing within-prompt variance. Additionally, our statistical model assumes that all prompts are independently sampled from the benchmark difficulty distribution, which may not be accurate in practice, as prompts can originate from the same subjects or resources. Future work should consider incorporating the covariance structure into the estimation process. Another drawback is the detection of mislabeled prompts. Although our method efficiently reduces the effort needed to filter samples, the true positive rate is not high (around $50\%$). Potential research could leverage more sophisticated semantic metrics and model ensembles to better detect mislabeled or ambiguous prompts.

\section*{Ethic Statement}
Our work utilizes benchmark datasets to evaluate LLMs. All the datasets and LLMs are publicly available.




\bibliography{ref}

@misc{open-llm-leaderboard-v1,
  author = {Edward Beeching and Clémentine Fourrier and Nathan Habib and Sheon Han and Nathan Lambert and Nazneen Rajani and Omar Sanseviero and Lewis Tunstall and Thomas Wolf},
  title = {Open LLM Leaderboard (2023-2024)},
  year = {2023},
  publisher = {Hugging Face},
}

@article{huang2024trustllm,
  title={Trustllm: Trustworthiness in large language models},
  author={Huang, Yue and Sun, Lichao and Wang, Haoran and Wu, Siyuan and Zhang, Qihui and Li, Yuan and Gao, Chujie and Huang, Yixin and Lyu, Wenhan and Zhang, Yixuan and others},
  journal={arXiv preprint arXiv:2401.05561},
  year={2024}
}

@article{wang2024mmlu,
  title={Mmlu-pro: A more robust and challenging multi-task language understanding benchmark},
  author={Wang, Yubo and Ma, Xueguang and Zhang, Ge and Ni, Yuansheng and Chandra, Abhranil and Guo, Shiguang and Ren, Weiming and Arulraj, Aaran and He, Xuan and Jiang, Ziyan and others},
  journal={arXiv preprint arXiv:2406.01574},
  year={2024}
}

@article{song2024good,
  title={The good, the bad, and the greedy: Evaluation of llms should not ignore non-determinism},
  author={Song, Yifan and Wang, Guoyin and Li, Sujian and Lin, Bill Yuchen},
  journal={arXiv preprint arXiv:2407.10457},
  year={2024}
}

@inproceedings{hendrycks2measuring,
  title={Measuring Mathematical Problem Solving With the MATH Dataset},
  author={Hendrycks, Dan and Burns, Collin and Kadavath, Saurav and Arora, Akul and Basart, Steven and Tang, Eric and Song, Dawn and Steinhardt, Jacob},
  booktitle={Thirty-fifth Conference on Neural Information Processing Systems Datasets and Benchmarks Track (Round 2)}
}

@article{rein2023gpqa,
  title={Gpqa: A graduate-level google-proof q\&a benchmark},
  author={Rein, David and Hou, Betty Li and Stickland, Asa Cooper and Petty, Jackson and Pang, Richard Yuanzhe and Dirani, Julien and Michael, Julian and Bowman, Samuel R},
  journal={arXiv preprint arXiv:2311.12022},
  year={2023}
}

@article{ding2024easy2hard,
  title={Easy2Hard-Bench: Standardized Difficulty Labels for Profiling LLM Performance and Generalization},
  author={Ding, Mucong and Deng, Chenghao and Choo, Jocelyn and Wu, Zichu and Agrawal, Aakriti and Schwarzschild, Avi and Zhou, Tianyi and Goldstein, Tom and Langford, John and Anandkumar, Anima and others},
  journal={arXiv preprint arXiv:2409.18433},
  year={2024}
}

@inproceedings{polotinybenchmarks,
  title={tinyBenchmarks: evaluating LLMs with fewer examples},
  author={Polo, Felipe Maia and Weber, Lucas and Choshen, Leshem and Sun, Yuekai and Xu, Gongjun and Yurochkin, Mikhail},
  booktitle={Forty-first International Conference on Machine Learning}
}

@article{glickman2012example,
  title={Example of the Glicko-2 system},
  author={Glickman, Mark E},
  journal={Boston University},
  volume={28},
  year={2012}
}

@article{cai2016item,
  title={Item response theory},
  author={Cai, Li and Choi, Kilchan and Hansen, Mark and Harrell, Lauren},
  journal={Annual Review of Statistics and Its Application},
  volume={3},
  number={1},
  pages={297--321},
  year={2016},
  publisher={Annual Reviews}
}

@article{natesan2016bayesian,
  title={Bayesian prior choice in IRT estimation using MCMC and variational Bayes},
  author={Natesan, Prathiba and Nandakumar, Ratna and Minka, Tom and Rubright, Jonathan D},
  journal={Frontiers in psychology},
  volume={7},
  pages={1422},
  year={2016},
  publisher={Frontiers Media SA}
}

@article{desender2017subjective,
  title={Subjective experience of difficulty depends on multiple cues},
  author={Desender, Kobe and Van Opstal, Filip and Van den Bussche, Eva},
  journal={Scientific reports},
  volume={7},
  number={1},
  pages={44222},
  year={2017},
  publisher={Nature Publishing Group UK London}
}

@article{hendrycks2020measuring,
  title={Measuring massive multitask language understanding},
  author={Hendrycks, Dan and Burns, Collin and Basart, Steven and Zou, Andy and Mazeika, Mantas and Song, Dawn and Steinhardt, Jacob},
  journal={arXiv preprint arXiv:2009.03300},
  year={2020}
}

@article{liang2022holistic,
  title={Holistic evaluation of language models},
  author={Liang, Percy and Bommasani, Rishi and Lee, Tony and Tsipras, Dimitris and Soylu, Dilara and Yasunaga, Michihiro and Zhang, Yian and Narayanan, Deepak and Wu, Yuhuai and Kumar, Ananya and others},
  journal={arXiv preprint arXiv:2211.09110},
  year={2022}
}

@article{srivastava2022beyond,
  title={Beyond the imitation game: Quantifying and extrapolating the capabilities of language models},
  author={Srivastava, Aarohi and Rastogi, Abhinav and Rao, Abhishek and Shoeb, Abu Awal Md and Abid, Abubakar and Fisch, Adam and Brown, Adam R and Santoro, Adam and Gupta, Aditya and Garriga-Alonso, Adri{\`a} and others},
  journal={arXiv preprint arXiv:2206.04615},
  year={2022}
}

@article{zhang2024defining,
  title={Defining Boundaries: A Spectrum of Task Feasibility for Large Language Models},
  author={Zhang, Wenbo and Xu, Zihang and Cai, Hengrui},
  journal={arXiv preprint arXiv:2408.05873},
  year={2024}
}

@article{cobbe2021training,
  title={Training verifiers to solve math word problems},
  author={Cobbe, Karl and Kosaraju, Vineet and Bavarian, Mohammad and Chen, Mark and Jun, Heewoo and Kaiser, Lukasz and Plappert, Matthias and Tworek, Jerry and Hilton, Jacob and Nakano, Reiichiro and others},
  journal={arXiv preprint arXiv:2110.14168},
  year={2021}
}

@article{sprague2023musr,
  title={Musr: Testing the limits of chain-of-thought with multistep soft reasoning},
  author={Sprague, Zayne and Ye, Xi and Bostrom, Kaj and Chaudhuri, Swarat and Durrett, Greg},
  journal={arXiv preprint arXiv:2310.16049},
  year={2023}
}

@article{zhou2023instruction,
  title={Instruction-following evaluation for large language models},
  author={Zhou, Jeffrey and Lu, Tianjian and Mishra, Swaroop and Brahma, Siddhartha and Basu, Sujoy and Luan, Yi and Zhou, Denny and Hou, Le},
  journal={arXiv preprint arXiv:2311.07911},
  year={2023}
}

@article{polo2024tinybenchmarks,
  title={tinyBenchmarks: evaluating LLMs with fewer examples},
  author={Polo, Felipe Maia and Weber, Lucas and Choshen, Leshem and Sun, Yuekai and Xu, Gongjun and Yurochkin, Mikhail},
  journal={arXiv preprint arXiv:2402.14992},
  year={2024}
}

@article{madaan2024quantifying,
  title={Quantifying Variance in Evaluation Benchmarks},
  author={Madaan, Lovish and Singh, Aaditya K and Schaeffer, Rylan and Poulton, Andrew and Koyejo, Sanmi and Stenetorp, Pontus and Narang, Sharan and Hupkes, Dieuwke},
  journal={arXiv preprint arXiv:2406.10229},
  year={2024}
}

@article{dubey2024llama,
  title={The llama 3 herd of models},
  author={Dubey, Abhimanyu and Jauhri, Abhinav and Pandey, Abhinav and Kadian, Abhishek and Al-Dahle, Ahmad and Letman, Aiesha and Mathur, Akhil and Schelten, Alan and Yang, Amy and Fan, Angela and others},
  journal={arXiv preprint arXiv:2407.21783},
  year={2024}
}

@article{qwen2,
      title={Qwen2 Technical Report}, 
      author={An Yang and Baosong Yang and Binyuan Hui and Bo Zheng and Bowen Yu and Chang Zhou and Chengpeng Li and Chengyuan Li and Dayiheng Liu and Fei Huang and Guanting Dong and Haoran Wei and Huan Lin and Jialong Tang and Jialin Wang and Jian Yang and Jianhong Tu and Jianwei Zhang and Jianxin Ma and Jin Xu and Jingren Zhou and Jinze Bai and Jinzheng He and Junyang Lin and Kai Dang and Keming Lu and Keqin Chen and Kexin Yang and Mei Li and Mingfeng Xue and Na Ni and Pei Zhang and Peng Wang and Ru Peng and Rui Men and Ruize Gao and Runji Lin and Shijie Wang and Shuai Bai and Sinan Tan and Tianhang Zhu and Tianhao Li and Tianyu Liu and Wenbin Ge and Xiaodong Deng and Xiaohuan Zhou and Xingzhang Ren and Xinyu Zhang and Xipin Wei and Xuancheng Ren and Yang Fan and Yang Yao and Yichang Zhang and Yu Wan and Yunfei Chu and Yuqiong Liu and Zeyu Cui and Zhenru Zhang and Zhihao Fan},
      journal={arXiv preprint arXiv:2407.10671},
      year={2024}
}

@article{jiang2023mistral,
  title={Mistral 7B},
  author={Jiang, Albert Q and Sablayrolles, Alexandre and Mensch, Arthur and Bamford, Chris and Chaplot, Devendra Singh and Casas, Diego de las and Bressand, Florian and Lengyel, Gianna and Lample, Guillaume and Saulnier, Lucile and others},
  journal={arXiv preprint arXiv:2310.06825},
  year={2023}
}

@article{white2024livebench,
  title={Livebench: A challenging, contamination-free llm benchmark},
  author={White, Colin and Dooley, Samuel and Roberts, Manley and Pal, Arka and Feuer, Ben and Jain, Siddhartha and Shwartz-Ziv, Ravid and Jain, Neel and Saifullah, Khalid and Naidu, Siddartha and others},
  journal={arXiv preprint arXiv:2406.19314},
  year={2024}
}

@article{lin2024wildbench,
  title={WILDBENCH: Benchmarking LLMs with Challenging Tasks from Real Users in the Wild},
  author={Lin, Bill Yuchen and Deng, Yuntian and Chandu, Khyathi and Brahman, Faeze and Ravichander, Abhilasha and Pyatkin, Valentina and Dziri, Nouha and Bras, Ronan Le and Choi, Yejin},
  journal={arXiv preprint arXiv:2406.04770},
  year={2024}
}

@article{zheng2023judging,
  title={Judging llm-as-a-judge with mt-bench and chatbot arena},
  author={Zheng, Lianmin and Chiang, Wei-Lin and Sheng, Ying and Zhuang, Siyuan and Wu, Zhanghao and Zhuang, Yonghao and Lin, Zi and Li, Zhuohan and Li, Dacheng and Xing, Eric and others},
  journal={Advances in Neural Information Processing Systems},
  volume={36},
  pages={46595--46623},
  year={2023}
}

@misc{alpaca_eval,
  author = {Xuechen Li and Tianyi Zhang and Yann Dubois and Rohan Taori and Ishaan Gulrajani and Carlos Guestrin and Percy Liang and Tatsunori B. Hashimoto },
  title = {AlpacaEval: An Automatic Evaluator of Instruction-following Models},
  year = {2023},
  month = {5},
  publisher = {GitHub},
  journal = {GitHub repository},
  howpublished = {\url{https://github.com/tatsu-lab/alpaca_eval}}
}

@article{holtzman2019curious,
  title={The curious case of neural text degeneration},
  author={Holtzman, Ari and Buys, Jan and Du, Li and Forbes, Maxwell and Choi, Yejin},
  journal={arXiv preprint arXiv:1904.09751},
  year={2019}
}

@article{wang2022self,
  title={Self-consistency improves chain of thought reasoning in language models},
  author={Wang, Xuezhi and Wei, Jason and Schuurmans, Dale and Le, Quoc and Chi, Ed and Narang, Sharan and Chowdhery, Aakanksha and Zhou, Denny},
  journal={arXiv preprint arXiv:2203.11171},
  year={2022}
}

@inproceedings{mitchell2022enhancing,
  title={Enhancing Self-Consistency and Performance of Pre-Trained Language Models through Natural Language Inference},
  author={Mitchell, Eric and Noh, Joseph and Li, Siyan and Armstrong, Will and Agarwal, Ananth and Liu, Patrick and Finn, Chelsea and Manning, Christopher D},
  booktitle={Proceedings of the 2022 Conference on Empirical Methods in Natural Language Processing},
  pages={1754--1768},
  year={2022}
}

@article{swayamdipta2020dataset,
  title={Dataset cartography: Mapping and diagnosing datasets with training dynamics},
  author={Swayamdipta, Swabha and Schwartz, Roy and Lourie, Nicholas and Wang, Yizhong and Hajishirzi, Hannaneh and Smith, Noah A and Choi, Yejin},
  journal={arXiv preprint arXiv:2009.10795},
  year={2020}
}

@article{miller2024adding,
  title={Adding Error Bars to Evals: A Statistical Approach to Language Model Evaluations},
  author={Miller, Evan},
  journal={arXiv preprint arXiv:2411.00640},
  year={2024}
}

@inproceedings{wang2023decodingtrust,
  title={DecodingTrust: A Comprehensive Assessment of Trustworthiness in GPT Models.},
  author={Wang, Boxin and Chen, Weixin and Pei, Hengzhi and Xie, Chulin and Kang, Mintong and Zhang, Chenhui and Xu, Chejian and Xiong, Zidi and Dutta, Ritik and Schaeffer, Rylan and others},
  booktitle={NeurIPS},
  year={2023}
}

\clearpage
\appendix
\onecolumn

\section{IRT is a special parametrization of $\mathbb P\left(\text{correct}\right)$}
\label{irt section}
 $\mathbb P\left(\text{correct}\right)$ is closely connected to item response theory. Many studies \citep{polo2024tinybenchmarks,madaan2024quantifying,ding2024easy2hard} utilize IRT to quantify the difficulty of prompts using multiple LLMs. One variation of the IRT model is the one-parameter logistic (1PL) model as defined below: 
 \begin{align}
     \mathbb{P}\left(y_{li}=1\mid \theta_l,b_i\right) = \frac{1}{1+\exp^{-\left(-\theta_l-b_i\right)}},
     \label{irt}
 \end{align}
 where $\mathbb{P}\left(y_{li}=1\mid \theta_l,b_i\right)$ is the probability that LLM $l$ can answer the $j$-th prompt correctly. $\theta_l$ represents the latent ability of LLM $l$, $b_i$ is the difficulty parameter of the $j$-th prompt.

We observe that when we focus on a single LLM, i.e., when LLM $l$ is fixed, $\mathbb{P}\left(y_{li}=1\mid \theta_l,b_i\right)$ coincides with the prompt difficulty $p_i$ defined in (\ref{stat model}). Consequently, the right-hand side of (\ref{irt}) can be viewed as a specific parametrization of the prompt difficulty using a logit link function. This implies that, theoretically, the maximum likelihood estimator of IRT and our method are equivalent via a sigmoid transformation. We use the 1PL model here for illustrative purposes, but this equivalence also holds when extended to models with more parameters.



\section{Benchmark Details}
\label{bench details}
MMLU-Pro is a comprehensive benchmark tailored for advanced, multi-disciplinary language understanding and reasoning at the proficient level. The GSM8K dataset comprises linguistically diverse math word problems from grade school curricula, crafted by human experts. MuSR is a specialized dataset designed to assess language models' performance on multi-step soft reasoning tasks presented in natural language narratives. IFEval, meanwhile, provides verifiable instructions to test large language models' ability to follow instructions accurately. 

\section{Additional Results on Varying Temperature $T$}
\label{vary T}
\begin{figure}[!htbp]
    \centering
    \subfigure[GSM8K Llama 8B]{\includegraphics[width=0.48\textwidth]{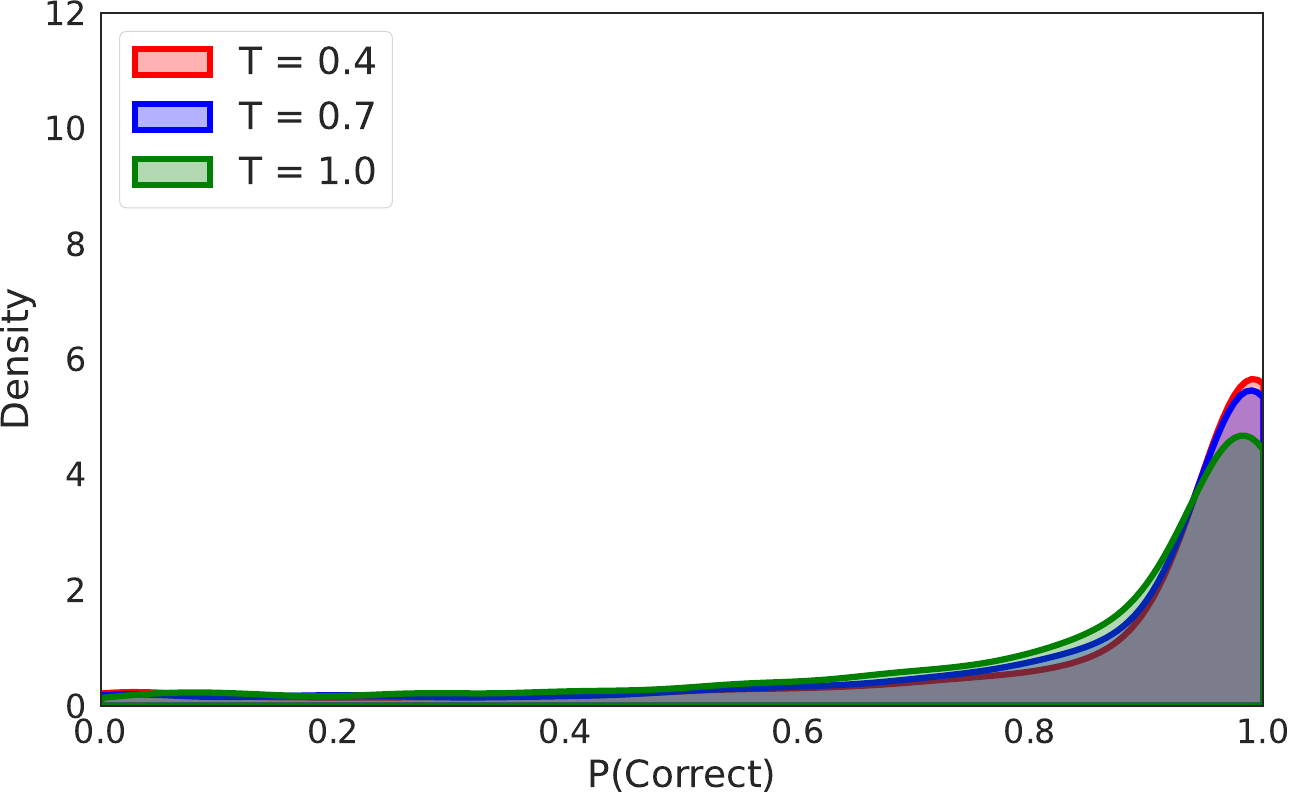}}
    \subfigure[GSM8K Llama 70B]
    {\includegraphics[width=0.48\textwidth]{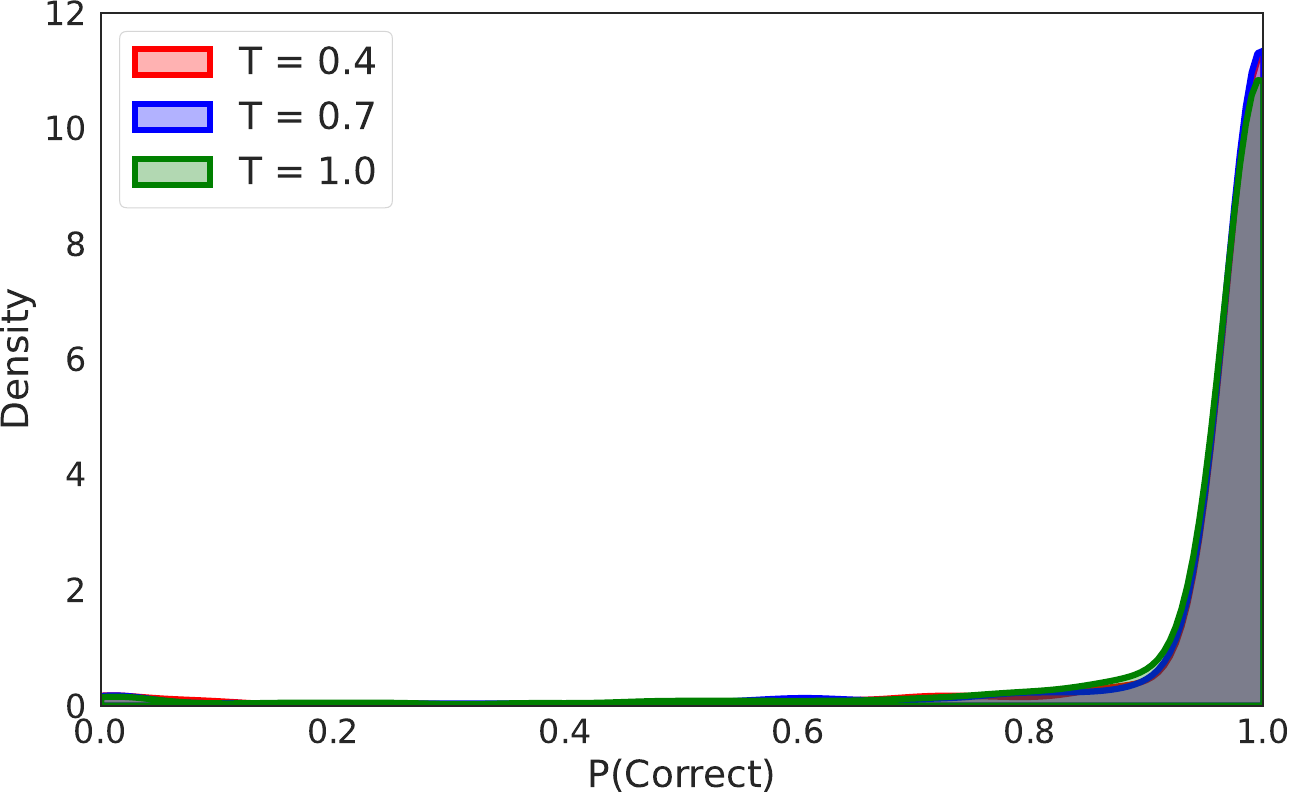}}
    \subfigure[MUSR Llama 8B]{\includegraphics[width=0.48\textwidth]{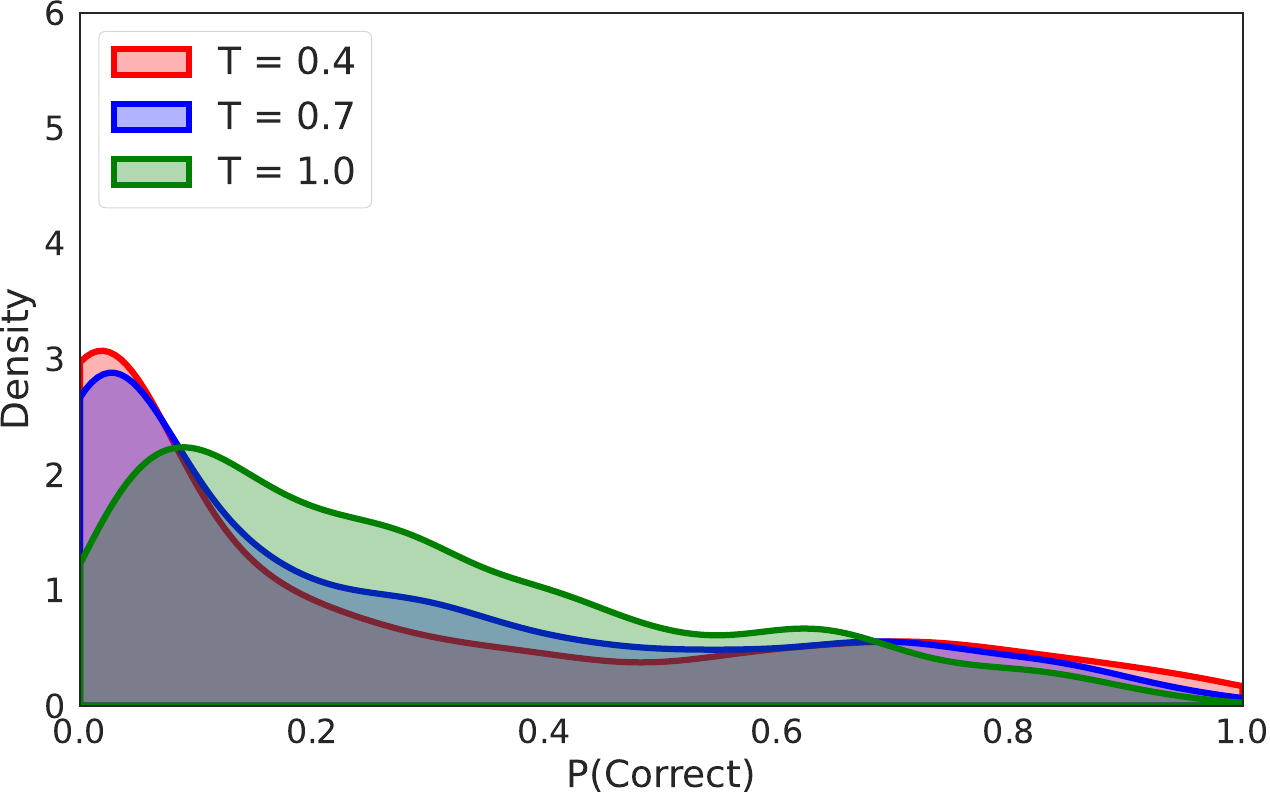}}
    \subfigure[MUSR Llama 70B]
    {\includegraphics[width=0.48\textwidth]{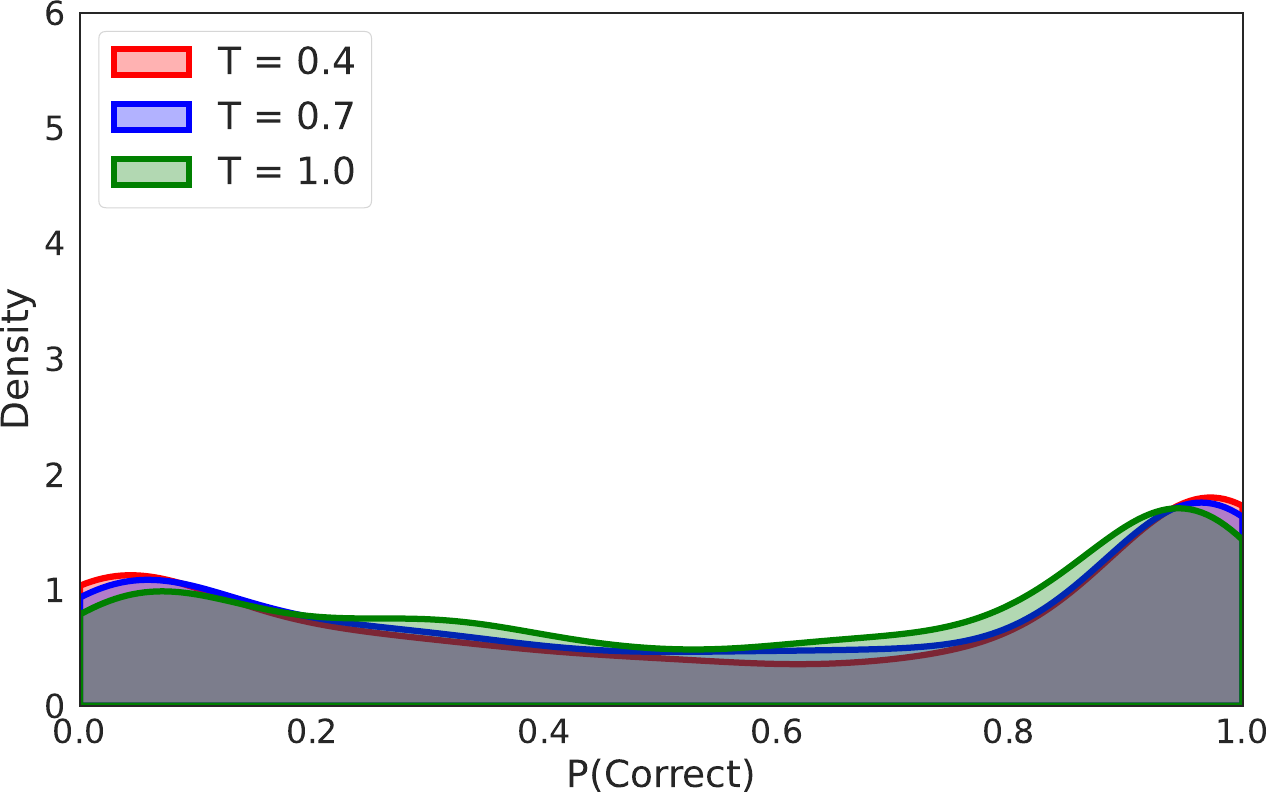}}
    \\
    \caption{Distribution of $\mathbb P\left(\text{correct}\right)$ for GSM8K and MUSR when varying temperature $T$.}
    \label{fig vary t}
\end{figure}

To investigate how temperature influences the $\mathbb{P}(\text{correct})$ distribution, we vary the sampling temperatures $T$ across $0.4$, $0.7$, and $1.0$ for the GSM8K and MUSR datasets using the Llama 8B and 70B models. The results are in Figure \ref{vary T}. We find that for the smaller 8B model, as $T$ increases, the distribution becomes more unstable with a more diffuse density. However, for the larger model, the $\mathbb{P}(\text{correct})$ is less sensitive to changes in $T$.

\section{Semantic Consistency for Responses: $\mathbb S\left(\text{consistency}\right)$}
\label{s_consis}
Apart from the correctness, we can also measure the difficulty of benchmark prompts by examining the semantic complexity from
multiple generations. This is because analyzing the nature of errors produced by LLMs can provide valuable insights into their decision-making processes. Specifically, it can help us determine whether LLMs tend to make consistent or varied mistakes, shedding light on their limitations and potential areas for improvement. 

We can group responses into multiple clusters based on their semantic meaning using bidirectional entailment predictions from a Natural Language Inference (NLI) model, such as DeBERTa or a prompted large language model (LLM). 


One common metric for quantifying consistency is the number of semantic sets, originally developed for uncertainty quantification in LLMs. The number of semantic sets assumes that a higher number of distinct semantic sets corresponds to lower consistency. 

However, the number of semantic sets only considers the number of clusters, without taking into account the proportion of generations within each cluster. For instance, consider two scenarios with 8 generations and 2 clusters: one where 1 generation falls into the first cluster and 7 into the second, versus another where 4 generations fall into each cluster. While these scenarios clearly represent different levels of consistency, the semantic set metric fails to distinguish between them, highlighting the need for a more nuanced approach to evaluating consistency.

Here we utilize a metric called semantic set entropy to better account for the proportions of semantic clusters. Given a set of $k$ generations and cluster them into $C$ semantic sets, semantic set entropy can be represented as:
$$
\mathbb S\left(\text{consistency}\right) = \sum_{c=1}^C \text{Prop}_c\log \text{Prop}_c,
$$
where $\text{Prop}_c$ measures the proportion of generations in group $c$ and its empirical estimator $\widehat{\text{Prop}}_c=\frac{\text{\# generations in set c}}{m}$ with finite $m$ samples. This can be seen as negative semantic set entropy, the larger, more consistent.

\section{Influence on Model Ranking: an Illustrative Example}
 We demonstrate the benefits of using multiple generations for ranking through both empirical results and theoretical analysis. Here we use two LLMs as illustrations, but this analysis can be generalized to multiple LLMs.

For empirical results, we evaluated the challenging GPQA dataset using two models: Llama3.1-8B and Mistral-8B-Instruct-2410. In practice, when using multiple generations, Mistral-8B-Instruct-2410 consistently outperforms Llama3.1-8B across repeated trials. However, if only a single generation is used, there is a $20 \%$ chance that Llama3.1-8B appears to rank higher, introducing ranking errors when comparing models. For theoretical analysis, our theoretical framework can also be extended to this scenario. Specifically, as illustrated by $\operatorname{Pr}\left(\hat{\mu}_1>\hat{\mu}_2\right)=\Phi\left(\frac{\mu_1-\mu_2}{\sqrt{O\left(\frac{1}{n k}\right)+O\left(\frac{1}{n}\right)}}\right)$ where we assume the true benchmark scores satisfy $\mu_1>\mu_2$ and $\Phi$ is the CDF of the Gaussian Distribution. This expression shows how variance reduction from additional generations directly improves ranking reliability.
\section{Proof of Lemma \ref{lemma}}
Restate of Lemma \ref{lemma}: 

\noindent Given the model \begin{equation}
\begin{gathered}
p_i \sim \mathbb{P}(\mu, \sigma;\theta) \quad \text{for } i=1,\cdots,n\\
y_{i,j}  \sim \text{Bernoulli}(p_i) \quad \text{for } j=1,\cdots,k,
\end{gathered}
\end{equation}
and the moment estimator $\hat \mu =\frac{\sum_{i=1}^n\sum_{j=1}^k y_{i,j}}{nk}.$
Then $\hat \mu$ is an unbiased estimator for $\mu$ and its variance equals
    \begin{equation*}
\text{Var}(\hat\mu)=\underbrace{\frac{1}{nk}\left(\mu-\mu^2-\sigma^2\right)}_{\text{Withth-prompt Variance}}+\underbrace{\frac{1}{n}\sigma^2}_{\text{Between-prompt Variance}}.
    \end{equation*}

\noindent \textit{Proof}:
Firstly we show $\hat\mu$ is an unbiased estimation of $\mu$, which can be directly show by the expectation:

\begin{align*}
\mathbb{E}\left[\hat \mu\right] &= \frac{\sum_{i=1}^n\sum_{j=1}^k y_{i,j}}{nk} \\
& = \frac{\sum_{i=1}^n\mathbb E\left[\sum_{j=1}^k y_{i,j}\right]}{nk}\\
&  \stackrel{(3)}{=}  \frac{\sum_{i=1}^n\mathbb E\left[\mathbb E\left[\sum_{j=1}^k y_{i,j}\mid p_i\right]\right]}{nk} \\
& = \frac{\sum_{i=1}^n k\mathbb E\left[p_i\right]}{nk}\\
& = \frac{\sum_{i=1}^n k\mathbb \mu}{nk}\\
& = \mu, 
\end{align*}

where $(3)$ utilizes the law of total expectation. Hence $\hat \mu$ is unbiased estimator of $\mu$.
The variance of $\hat \mu$ can be further shown:

\begin{align*}
\text{Var}\left(\hat\mu\right)&=\text{Var}\left(\frac{\sum_{i=1}^n\sum_{j=1}^k y_{i,j}}{nk}\right)\\
& = \frac{1}{n^2k^2 }\left(\sum_{i=1}^n\text{Var}\left(\sum_{j=1}^k y_{i,j}\right)\right)\\
& \stackrel{(3)}{=} \frac{1}{n^2k^2}\left(\sum_{i=1}^n \mathbb E \left[\text{Var}\left(\sum_{j=1}^k y_{ij}\mid p_i\right)\right]\right.\\ 
&\quad + \left. \text{Var} \left(\mathbb E \left(\sum_{j=1}^k y_{ij}\mid p_i\right)\right)\right)\\
& = \frac{1}{n^2k^2}\left(\sum_{i=1}^n \mathbb E\left[ k p_i\left(1-p_i\right)\right] + \text{Var}\left(kp_i\right) \right)\\
& = \frac{1}{n^2k^2} \left(nk \left(\mathbb E\left[p_i\right] - \mathbb E\left[p_i^2\right]\right)+ nk^2\text{Var}\left(p_i\right)\right)\\
&=\underbrace{\frac{1}{nk}\left(\mu-\mu^2-\sigma^2\right)}_{\text{Withth-prompt Variance}}+\underbrace{\frac{1}{n}\sigma^2}_{\text{Between-prompt Variance}}.
\end{align*}
where $(3)$ utilizes the low of total variance.


\begin{figure}[!htbp]
    \centering
    \includegraphics[width=\columnwidth]{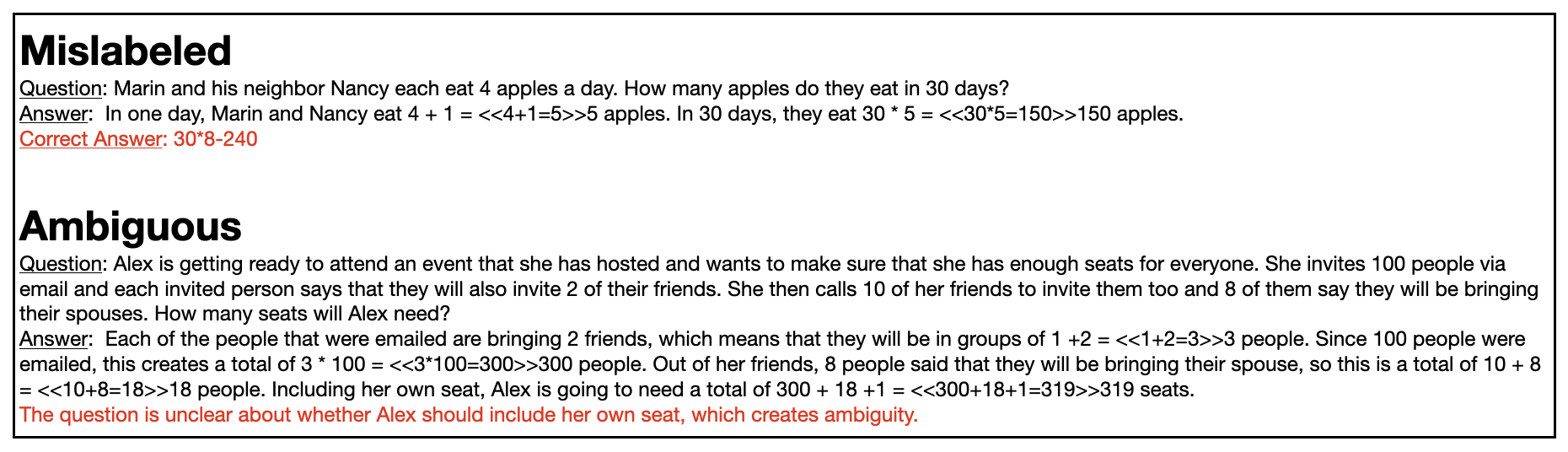}
    \caption{Examples of detected mislabeled and ambiguous prompts in GSM8K.}
    \label{example}
\end{figure}

\end{document}